\documentclass[lettersize,journal]{IEEEtran}
\usepackage{amsmath,amsfonts}
\usepackage{enumitem}
\usepackage{algorithmic}
\usepackage{algorithm}
\usepackage{array}
\usepackage[caption=false,font=normalsize,labelfont=sf,textfont=sf]{subfig}
\usepackage{textcomp}
\usepackage{stfloats}
\usepackage{url}
\usepackage{verbatim}
\usepackage{graphicx}
\usepackage{multirow}
\usepackage{color}
\usepackage{todonotes}

\usetikzlibrary{positioning}
\usepackage[square, numbers, sort&compress]{natbib}
\begin{document}

\title{Learning to Hop for a Single-Legged Robot with Parallel Mechanism}

\author{Hongbo Zhang$^1$, Xiangyu Chu$^{1,2}$, Yanlin Chen$^{1,2}$, Yunxi Tang$^{1}$, Linzhu Yue$^1$, \\ Yun-Hui Liu$^1$, and Kwok Wai Samuel Au$^{1,2}$
\thanks{
$^{1}$Department of Mechanical and Automation Engineering, The Chinese University of Hong Kong, Hong Kong SAR, China. $^{2}$Multiscale Medical Robotics Centre

E-mail: hbzhang@mae.cuhk.edu.hk



}
}



\maketitle

\begin{abstract}
This work presents the application of reinforcement learning to improve the performance of a highly dynamic hopping system with a parallel mechanism. Unlike serial mechanisms, parallel mechanisms can not be accurately simulated due to the complexity of their kinematic constraints and closed-loop structures. Besides, learning to hop suffers from prolonged aerial phase and the sparse nature of the rewards. To address them, we propose a learning framework to encode long-history feedback to account for the under-actuation brought by the prolonged aerial phase. In the proposed framework, we also introduce a simplified serial configuration for the parallel design to avoid directly simulating parallel structure during the training. A torque-level conversion is designed to deal with the parallel-serial conversion to handle the sim-to-real issue. Simulation and hardware experiments have been conducted to validate this framework.
\end{abstract}


\section{Introduction}

   Legged robots have demonstrated significant potential for navigating complex environments, with hopping emerging as a particularly effective locomotion strategy for overcoming obstacles and traversing uneven terrain. Inspired by the agility and traversal capabilities of animals~\cite{burrows2006jumping}, single-legged robots represent a natural and minimalist design. Their relatively lower degrees of freedom (DoF) make them well-suited for emulating natural movements, enabling applications in exploration, search-and-rescue, and inspection tasks. Despite advancements in single-legged hopping control reported in the literature~\cite{kalouche2017goat,batts2017untethered,haldane2016robotic}, current methods still fall short of achieving the mobility required for real-world applications. The simplicity of a single-legged robot’s structure contrasts greatly with the complexity of its hopping control, which is particularly challenging due to extended aerial phases. The robot relies solely on adjustments made in the stance phase, using a single foot to recover from disturbances and prepare for subsequent hops. In contrast to jumping motions, which emphasize explosive power, hopping motions demand continuity and stability, underscoring the need for more robust control strategies to enhance hopping performance in practical scenarios.


\begin{figure}[!t]
\centering
\includegraphics[width=0.99\linewidth]{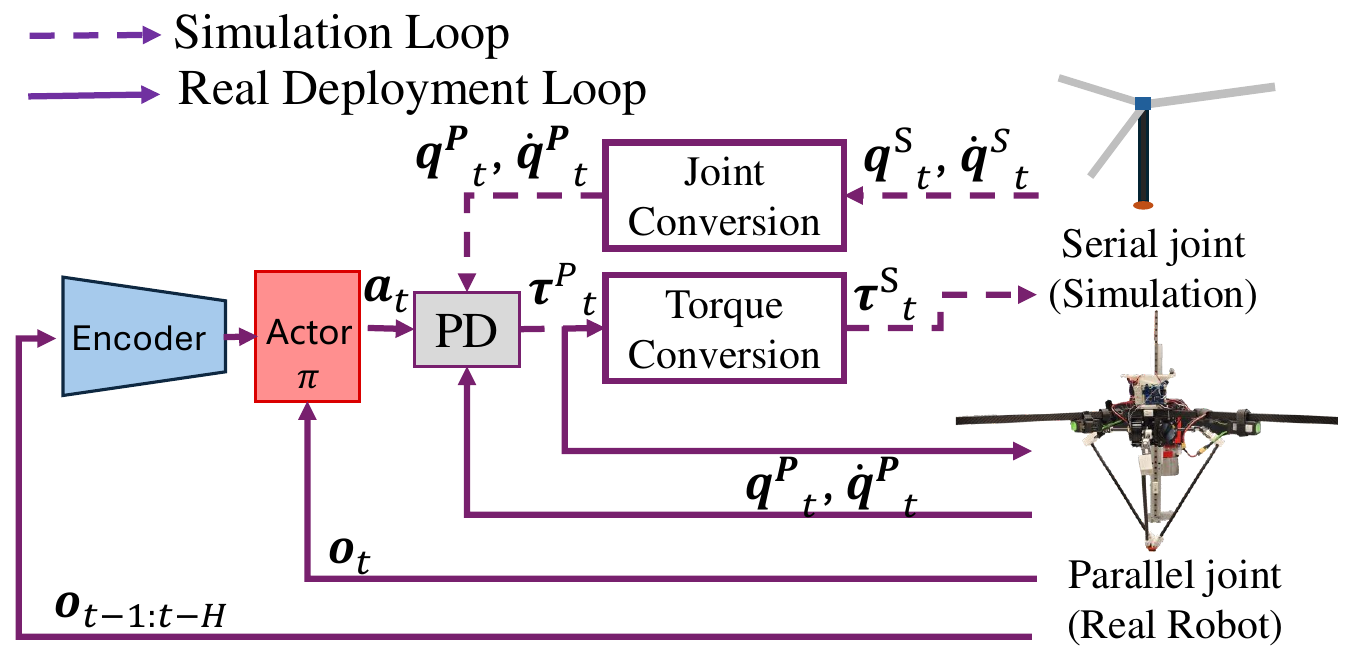}
\vspace{-0.3cm}
\caption{Overview of the hopping control framework for the single-legged robot with a parallel mechanism using reinforcement learning.}
\label{overview}
\end{figure}

Hopping control typically consists of two distinct phases. During the flight phase, a proportional-derivative (PD) controller is employed to achieve the desired leg angle for accurate landing. In the stance phase, a torque or force controller is used to maintain body orientation and regulate the take-off velocity. Existing approaches to hopping control have been continuously inspired by the pioneering work of Raibert \cite{RaibertHoppingControl}, which demonstrated the feasibility of using heuristic controllers with a decomposed control architecture to enable highly dynamic legged locomotion. Subsequently, numerous studies on single-legged robot control have been conducted, such as \cite{NMPC_hopping}, \cite{HoppingControl_Evangelos}, and \cite{NMPC_hopping_S2S}. While effective in generating stable hopping behavior, these controllers often require extensive parameter tuning and are prone to model uncertainties.



The recent emergence of reinforcement learning (RL) methods has opened new possibilities for developing  stable and robust locomotion controllers for legged robotic systems~\cite{lee2020learning,li2024reinforcement}. RL enables robots to learn complex behaviors through trial and error, optimizing their movements based on feedback from the environment. Promising results have been reported on learning to jump on bipedal and quadrupedal systems~\cite{li2023robust,yang2023continuous}. For these multi-legged systems, the challenges in designing jumping motions arise mainly from their high degrees of freedom (DoF), requiring careful coordination among multiple joints. On the other hand, continuous hopping for single-legged robotic systems poses different challenges, primarily due to the small supporting polygon and extended aerial phases, which demand highly precise control. While there have been studies applying RL to single-legged robots, these approaches often do not fully address the unique difficulties posed by such systems. For instance, \cite{kawaharazuka2022continuous} proposed a method for achieving continuous hopping with a parallel wire-driven monopedal robot, and \cite{RL_hopping} introduced an end-to-end RL approach for a monopedal robot with articulated joints. However, these works did not explicitly tackle the inherent challenges of single-legged robots within their RL frameworks.

The canonical system for studying hopping is single-legged robots, most of which use serial mechanism-based legs. Serial mechanisms are widely used, and they are relatively easier to simulate and optimize in physics engines while developing learning-based methods. In contrast, parallel mechanism-based legs remain underexplored, despite offering significant advantages. Parallel mechanisms provide higher stiffness, better load distribution, and improved force transmission efficiency, making them well-suited for robust and efficient hopping performance \cite{GOAT}. However, the use of parallel mechanism-based legs introduces substantial challenges for developing learning-based hopping controllers. A key difficulty lies in the inability of existing physics engines to accurately simulate the coupled dynamics and kinematic constraints of parallel mechanisms. This lack of accurate physical simulation prevents effective policy training in simulators. These challenges also arise on other kinds of robots containing parallel designs such as some humanoids with parallel designed legs~\cite{li2024reinforcement,SR-digit,gu2024advancing}. In these works, either some assumptions and simplifications have been made to handle the parallel design in simulation or kinematics conversion has been made to complete the sim-to-real transfer. In this work, we hypothesize that the 3D parallel design of this hopping robot makes the sim-to-real transfer even harder, and simple simplification or kinematics conversion fails to work in our case. We design a novel torque level conversion to handle the sim-to-real transfer problem.

In this work, we aim to design a continuous hopping controller for a single-legged robot with a parallel mechanism-based leg, by employing reinforcement learning based on an equivalent serial mechanism and torque-level conversion between serial and parallel mechanisms, as shown in Fig.~\ref{overview}. The proposed method can alleviate the under-actuation nature of the hopper system by introducing explicit velocity from a long history of proprioception. A hopping robot with a parallel 3-RSR design is employed to test the proposed hopping controller. Simulations and real-world experiments have verified the design of this learning-based controller. Also, comparisons have been made to verify the effectiveness of our proposed sim-to-real conversion method.

The primary contributions of this work are:
\begin{enumerate}
    \item We propose an RL-based control that solves the continuous hopping control for a single-legged robot system.
    \item A new torque-level conversion method is proposed to solve the sim-to-real issue brought by the 3D parallel joint configuration of the robot.
    \item Systematic simulations and experiments validate the feasibility and effectiveness of the proposed learning-based controller. 
\end{enumerate}


\section{Platform Overview}
\subsection{Hopping Robot Design}
\begin{figure}[!t]
\centering
\includegraphics[width=0.99\linewidth]{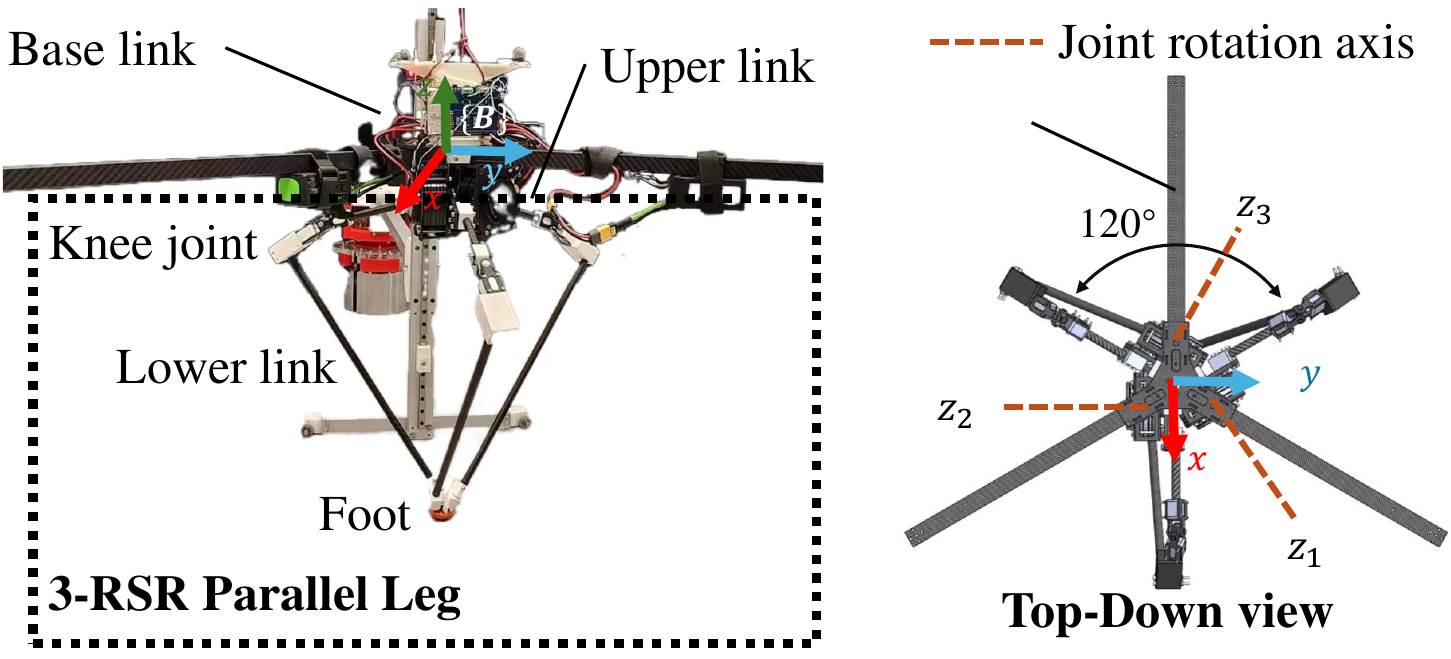}
\vspace{-0.4cm}
\caption{Design and major parts of the 3-RSR single-legged hopping robot. The robot foot has three degrees of translational freedom.}
\label{design}
\end{figure}

\begin{figure}[!t]
\centering
\includegraphics[width=0.99\linewidth]{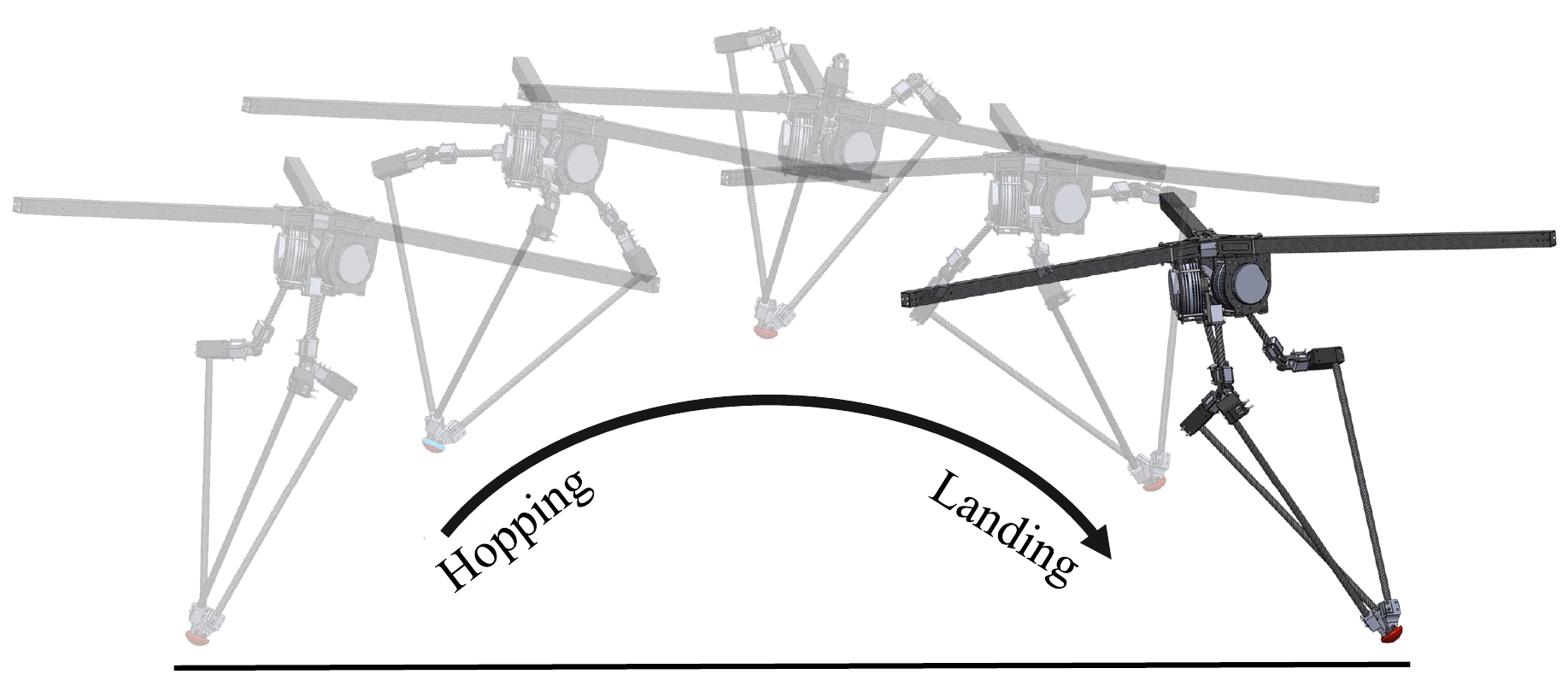}
\vspace{-0.4cm}
\caption{Hopping pattern of the 3D single-legged hopping robot.}
\label{jumping}
\end{figure}

The hopping robot features a main body with three balance rods and a 3-RSR parallel leg linkage with a point foot~\cite{sigma-hopper}, as shown in Fig.~\ref{design}. The main body comprises three identical geared motor assemblies, arranged at 120-degree angles relative to one another. The leg linkage is made up of three parallel chains, constructed from carbon fiber tubes, metal joints, and bearings. Driven by the hip motors, these chains provide three degrees of translational freedom (DoF) for the pointed foot. This robot's structure is designed to be minimal for 3D monopod hopping, and its hopping motion is illustrated in Fig.~\ref{jumping}. An open-source package for this hopping robot has been released\footnote{\url{https://github.com/CUHK-BRME/OMEGA}}.

The novel design of the single-legged robot brings some new challenges for the locomotion control problems, especially for the popular solution of legged locomotion control: reinforcement learning. We will elaborate on this from the following two aspects:
\subsection{Forward and Inverse Kinematics}
Assuming we have known the joint angles of three motor $\boldsymbol{q} = [q_1,q_2,q_3]$, the objective is to calculate the corresponding foot position $\mathbf{x}=[x_1,x_2,x_3]$ the position of the knee $\mathbf{k}_i$ can be geometrically represented  as:
\begin{equation}
\label{eq:knee}
\mathbf{k}_i = 
\begin{bmatrix}
k_{i1} \\
k_{i2} \\
k_{i3}
\end{bmatrix}
=
\mathbf{R}_i
\begin{bmatrix}
0 \\
r + D \cos(q_i) \\
D \sin(q_i)
\end{bmatrix},
\end{equation}
where \(i = 1, 2, 3\) represents each parallel chain and
$\mathbf{R}_i$ represents rotation for each parallel chain.
The combined 3RSR parallel leg design brings a geometric constrain: the distances from the foot to each knee are always kept the same:
\begin{equation}
\lvert \mathbf{x} - \mathbf{k}_1 \rvert^2 - \lvert \mathbf{x} - \mathbf{k}_2 \rvert^2 = \lvert \mathbf{x} - \mathbf{k}_3 \rvert^2 - \lvert \mathbf{x} - \mathbf{k}_2 \rvert^2 = 0,
\end{equation}
\begin{equation}
\lvert \mathbf{x} - \mathbf{k}_1 \rvert^2 = d^2.
\end{equation}

By rearranging the equations above, we can get the forward kinematics mapping $\textbf{FK}^{\mathcal{P}}$:
\begin{equation}
\label{constrains}
\mathbf{x} = \textbf{FK}^{\mathcal{P}}(\mathbf{q}) = 
\begin{bmatrix}
    A^{-1} B \\
    0
\end{bmatrix}
+ u 
\begin{bmatrix}
    A^{-1} C \\
    1
\end{bmatrix},
\end{equation}
where $A,B$ and $C$ matrix are:

\begin{equation*}
A = 
\left[
\begin{array}{cc}
    2(k_{11} - k_{21}) & 2(k_{12} - k_{22}) \\
    2(k_{31} - k_{21}) & 2(k_{32} - k_{22})
\end{array}
\right],
\end{equation*}
\begin{equation*}
\quad
B = 
\left[
\begin{array}{c}
    |k_1|^2 - |k_2|^2 \\
    |k_3|^2 - |k_2|^2
\end{array}
\right],
\quad
C = 
\left[
\begin{array}{c}
    -2(k_{13} - k_{23}) \\
    -2(k_{33} - k_{23})
\end{array}
\right].
\end{equation*}

The inverse kinematics is derived as follows. Assuming \( x \) is known and using Eq.~\ref{constrains}, we have
\begin{equation}
\label{ik_1}
(\mathbf{x} - \mathbf{k}_1)^T R_2 R_2^T (\mathbf{x} - \mathbf{k}_1) = d^2.
\end{equation}

After substituting Eq.~\ref{eq:knee} into Eq.~\ref{ik_1} and rearranging terms, the inverse kinematics for each joint can be written as:

\begin{equation}
q_i = \textbf{IK}^{\mathcal{P}}_{i}(x) = -\text{acos}\left( \frac{p_{ix}^2 + p_{iy}^2 + p_{iz}^2 + D^2 - d^2}{2D\sqrt{p_{iy}^2 + p_{iz}^2}} \right) + \alpha_i.
\end{equation}

Besides, the Jacobian of the parallel 3-RSR leg is calculated by numerically differentiating  $\textbf{IK}^{\mathcal{P}}(x)$ with auto differential tools such as pytorch~\cite{pytorch_NEURIPS2019_9015}:
\begin{equation}
\mathbf{J}^{\mathcal{P}} = \left( \frac{\partial IK}{\partial x} \right)^{-1} \in \mathbb{R}^{3 \times 3}.
\end{equation}

The readers can refer to~\cite{sigma-hopper} for elaborated kinematics derivation.
The kinematics and Jacobians of this hopping robot will be used for conversion between the parallel joint configuration and serial joint configuration.

\subsection{Parallel Mechanism}
\label{design:transfer}
Recently, researchers have reported extensively on handling the locomotion problem of legged robots with reinforcement learning. Different legged systems such as quadrupeds and bipedal have achieved satisfying performance on locomotion problems using reinforcement learning~\cite{SR-digit,lee2020learning}. In most work, policies are trained in simulation with accurately simulated rigid body dynamics and zero-shot transferred to the real robot. In these approaches, robots are designed to have serial layouts of legs which are easy to simulate. Our hopping robot, however, has 3D parallel leg dynamics which are hard to simulate accurately in most simulators. Extensive work including system modeling, system identification, and so on should be done to accurately simulate the real kinematics, dynamics, and contacts of the 3D parallel structure inside the simulator. 

To tackle this problem, we build a template model $\mathcal{S}$ that keeps the same Degree of Freedom (DoF) with the parallel design but has a serial joint configuration in the simulation. We train our locomotion policy based on the template model $\mathcal{S}$ in the simulation. During the real-world deployment, a joint torque mapping as shown in Eq.~\ref{Eq_torque_conversion} is performed to transfer the template joint torque into torques for the real joint configuration. 
This avoids directly training a policy on the original 3D paralleled structure which usually produces unreliable simulation results while also providing sufficient dynamic features of the real robot to facilitate sim-to-real transfer. Extensive real-world experiments have been conducted to verify the proposed sim-to-real methods.

\subsection{Underactuation of Continuous Hopping}
Creating stable and continuous hopping behaviour for a single-legged robot is challenging. During a hopping motion period, the hopping robot first makes hard contact with the ground and generates a large sudden change of force to push the robot upward while maintaining balance. A fully underactuated period (usually lasts for $0.2-0.4s$) will occur when the robot is in the air and free from any contact with the ground. After that, the robot should be prepared for landing and quickly adapt the ground reaction force to recover and power the next hopping motion. The whole agile hopping motion described above requires the controller to produce accurate force planning, predict landing contact after a long period of under-actuation and also maintain balance. To learn a hopping controller using reinforcement learning, we hypothesize that the difficulty lies in the handling of a long underactuated period, during which the robot's Center of Mass (CoM) is only influenced by gravity. The controller should learn to implicitly predict the landing and make adjustments accordingly. 

Addressing this challenge, we use an encoder-decoder structure~\cite{nahrendra2023dreamwaq} to extract hopping-related features and also generate an explicit estimation over the base velocity, especially the $z$ direction, from a history (0.1s) of proprioceptive measurements. Comparison results show that the latent vector extraction and the explicit estimation of the base linear velocity are crucial to producing a stable continuous hopping motion.

\section{Learning to Hop using Reinforcement Learning}
\begin{figure*}[!t]
\centering
\includegraphics[width=0.99\linewidth]{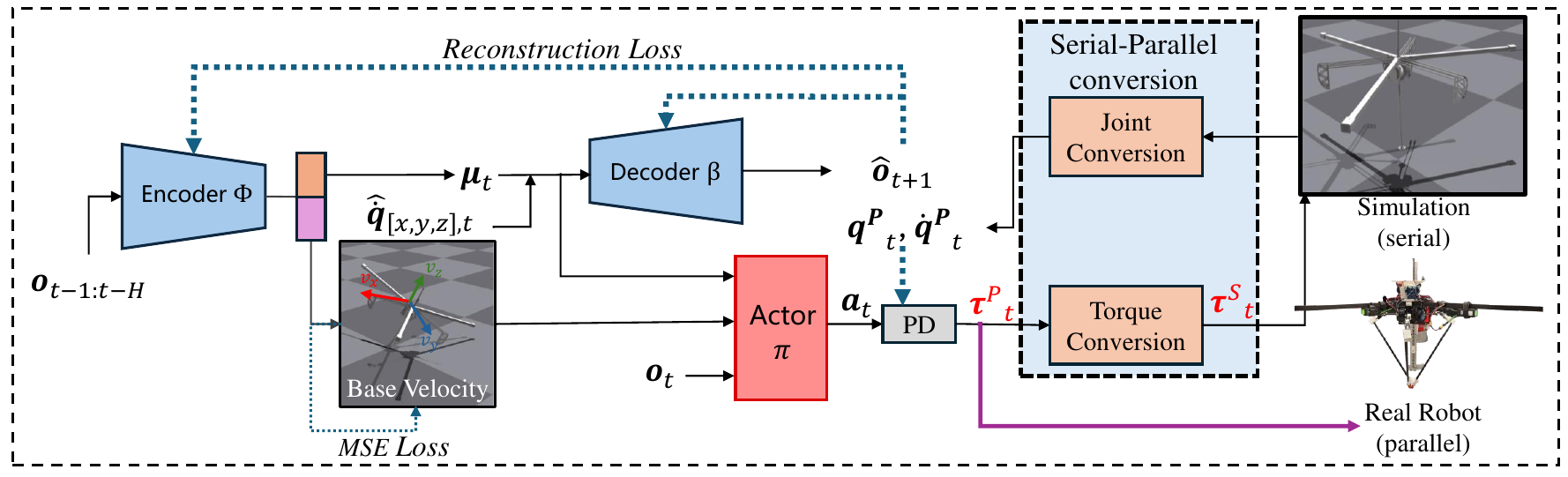}
\caption{\textbf{Actor policy structure of the proposed method.} The encoder $\phi$ receives a history ($H=5$) of observation and uses a multi-head structure to output a latent vector $\boldsymbol{\mu}_t$, and an estimation $(\dot{\mathbf{q}}_{[x,y,z],t},\mathbf{c}_t))$. The latent vector is then passed through the decoder $\beta$ to reconstruct a future observation $\mathbf{o}_{t+1}$. Following the design of $\beta$-VAE~\cite{higgins2017beta}, a reconstruction loss and a KL-divergence regulation loss are used to train the encoder-decoder. The actor receives the latent vector $\boldsymbol{\mu}_t$, current observation $\mathbf{o}_t$ and the estimated state $(\dot{\mathbf{q}}_{[x,y,z],t})$ to produce an action $\mathbf{a}_t$. 
A torque level conversion described in Eq.~\ref{Eq_torque_conversion} is done to map the serial joint torque to the parallel joint position.
}
\label{frame}
\end{figure*}

\subsection{Overview of the Framework}
We train a hopping policy for the parallel-designed hopping robot using reinforcement learning with the serial template model $\mathcal{S}$. Learning a balancing and velocity tracking controller for legged robots can be formulated as solving a POMDP which is well established in previous literatures~\cite{li2024reinforcement}. We adopt an asymmetric actor-critic structure and train the transferred template hopping robot as described in Sec.~\ref{design:transfer} in Isaac Gym~\cite{makoviychuk2021isaac}. The policy receives a history of proprioceptive history measurements while the critic has access to all the ground truth states. A Variational Auto Encoder (VAE)-style encoder-decoder structure is used to extract a compact representation from the history measurements. Also, base linear velocity \(\dot{\mathbf{q}}_{[x,y,z],t}\) is explicitly estimated through Mean Square Error (MSE) regression from the ground truth value.




\subsection{Reinforcement Learning Framework}
\label{method-learning}
We formulate this control problem as a POMDP. The definition of the Markov Decision Process can be described as follows:
\subsubsection{Observation Space}
The observation state of the robot $\mathbf{o}_t$ includes the parallel robot's measured motor positions $\hat{\mathbf{q}}^{\mathcal{P}}_{m,t}$, the base orientation $\hat{\mathbf{q}}_{[r,p,y],t}$, base angular velocity $\hat{\dot{\mathbf{q}}}_{[r,p,y],t}$, the $\cos$ and $\sin$ value of a phase timer $q_{\text{phase},t}$, the user command $\dot{\mathbf{q}}^{\text{d}}_{[x,y],t}$, desired gait period $T$ and previous action $\mathbf{a}^{\mathcal{P}}_{t-1}$:
\begin{align}
    \mathbf{o}_t = \big[ & \hat{\mathbf{q}}^{\mathcal{P}}_{m,t}, \hat{\mathbf{q}}_{[r,p,y],t}, \hat{\dot{\mathbf{q}}}_{[r,p,y],t}, 
    \cos(q_{\text{phase},t}), \sin(q_{\text{phase},t}), \notag \\
    & \dot{\mathbf{q}}^{\text{d}}_{[x,y],t}, T, \mathbf{a}_{t-1} \big].
\end{align}

Please note that the phase timer $q_{\text{phase},t}$ offers a periodical transition reference between the swing and stance phase for the hopper. During one swing-stance period, the value of $q_{\text{phase},t}$ linearly increases from $-2\pi$ to $2\pi$, where $q_{\text{phase},t}<0$ represents the stance phase and $q_{\text{phase},t}>0$ represents the swing phase.
The robot is encouraged to follow this schedule through the design of the reward. We found that adding a periodical signal to offer a fixed swing-stance schedule helps the agent generate better performance, especially for our morphology of single-legged dynamics. Also, the observation is chosen to directly receive the parallel joint motor position as input.


\subsubsection{Action Space}
The actor outputs the desired joint position $a_t$ under parallel configuration at 50Hz. A PD position controller deployed at 500Hz is then used to calculate the joint torques $\tau^{\mathcal{P}}$ from position tracking errors. In the simulation, the actual serial joint torque is mapped from it: $\tau^{\mathcal{S}} = g(\tau^{\mathcal{P}})$, where the calculation of the mapping $g(\cdot)$ is elaborated in Sec.~\ref{method-sim-to-real}. The $\tau^{\mathcal{S}}$ is then sent to the simulator. During the real-world deployment, no torque conversion is needed and $\tau^{\mathcal{P}}$ is sent directly to the real motor.

\subsubsection{Rewards}
The reward designs can be summarized into three major terms: tracking term, phase schedule term, and other auxiliary terms. Details can be found in the supplementary materials.

\begin{figure}[!t]
\centering
\includegraphics[width=0.99\linewidth]{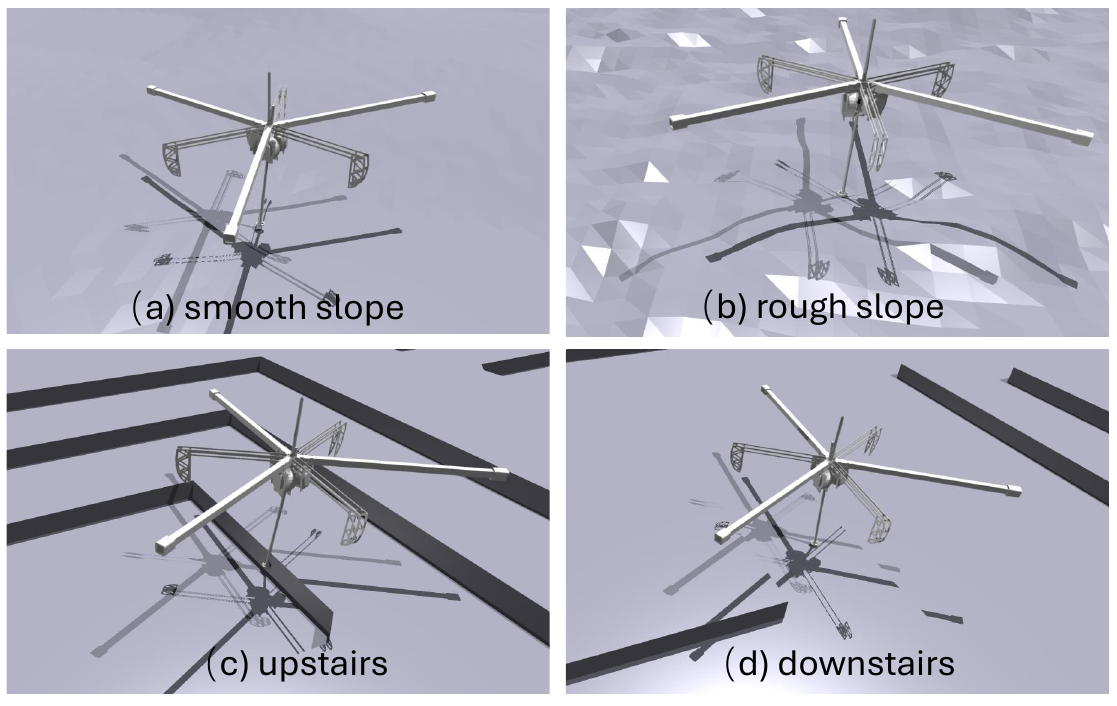}

\caption{ Training scenarios in the simulation for the single-legged robot. Four types of terrains including slopes and stairs are randomized and set up in the simulation. 
}
\label{simulation}
\vspace{-0.5cm}
\end{figure}

As shown in Fig.~\ref{frame}, the actor contains two parts of the network: an encoder-decoder structure $(\phi,\beta)$ and a base MLP policy $\pi$. The encoder $\phi$ receives a history length $H$ of observation: $\mathbf{o}_{t-1:t-H}$ and generates a latent vector $\boldsymbol{\mu}_t$, an estimation of base linear velocity \(\hat{\dot{\mathbf{q}}}_{[x,y,z],t}\). The latent vector $\boldsymbol{\mu}_t$ then passed through the decoder $\beta$ to reconstruct the future observation $\mathbf{o}_{t+1}$. Estimated base linear velocity \(\hat{\dot{\mathbf{q}}}_{[x,y,z],t}\) are fitted to the corresponding ground truth value obtained from the simulator. The latent vector $\boldsymbol{\mu}_t$ and the estimated value are passed to the base MLP $\pi(\mathbf{a}_t|\boldsymbol{\mu}_t,\hat{\dot{\mathbf{q}}}_{[x,y,z],t},\mathbf{o}_t)$ to generate the action $\mathbf{a}_t$.
For the critic, the full state including the actor observation and the ground truth states is injected into the critic represented by an MLP.

The training is done in Isaac Gym~\cite{makoviychuk2021isaac} using PPO~\cite{schulman2017proximal} with a template model with serial joint configuration as shown in Fig~\ref{frame}. We randomized several dynamic parameters and control parameters to increase the robustness of the hopping controller.
The transferred serial hopper model is trained in IsaacGym with a policy output frequency of $50$Hz. A PD controller is running at $200$Hz to convert the output position to joint torques. 

To produce a robust hopping policy and have a better sim-to-real performance, domain randomization over the robot dynamics is applied. Also, randomized terrains including smooth slopes, rough slopes, and stairs are set in the simulation as shown in Fig.~\ref{simulation}.



\subsection{Sim-to-Real Transfer}
\label{method-sim-to-real}
Due to the difficulty of accurately simulating the 3D parallel hopper robot in the simulation, we choose to convert the parallel joint configuration into the serial join configuration during the simulation, as shown in Fig.~\ref{frame}. We build a template model $\mathcal{S}$ with serial joint configuration in the simulation to approximate the kinematics and dynamics of the parallel design of the real robot. The template model also has 3 degrees of freedom and shares a similar end-effector workspace as the real robot. Three parallel joints are replaced by three serial joints including two perpendicular rotation joints and a prismatic joint as shown in Fig.~\ref{frame}. We train the hopping policy using this template model $\mathcal{S}$ in the simulation. This conversion keeps most part of the moving pattern of the original parallel model $\mathcal{P}$ but uses the serial joint configuration.

During the training, the policy is kept to be agnostic to the conversion. the policy $\pi$ receives a converted parallel joint position $\mathbf{q}^{\mathcal{P}}$ as shown in Eq.~\ref{Eq_torque_conversion} (a). The policy $\pi$ then directly outputs a desired joint position with parallel configuration $\mathbf{a}_t$. The torque $\boldsymbol{\tau}^{\mathcal{P}}$ is then calculated through a PD controller at 500 Hz as shown in Eq.~\ref{Eq_torque_conversion} (c).
A dynamic joint torque mapping ${\boldsymbol{\tau}^{\mathcal{P}} \rightarrow \boldsymbol{\tau}^{\mathcal{S}}}$ is applied to calculate the motor torque with serial joint configuration and sent to the simulator for calculating the dynamics of the serial model $\mathcal{S}$. As shown in Eq.~\ref{Eq_torque_conversion} (d), the conversion is done through Jacobian mapping with statics mechanics analysis.
\begin{subequations}
\label{Eq_torque_conversion}
\begin{align}
    \mathbf{q}^{\mathcal{P}} &= {\textbf {IK}}^{\mathcal{P}}({\textbf {FK}}^{\mathcal{S}}(\mathbf{q}^\mathcal{S})) \\
    \mathbf{\dot{q}}^{\mathcal{P}} &= (\mathbf{J}^{\mathcal{P}})^{-1}(\mathbf{J}^{\mathcal{S}}\mathbf{\dot{q}}^{\mathcal{S}})\\
    \boldsymbol{\tau}^{\mathcal{P}} &= \mathbf{K}_{\text p}(\mathbf{a}_t-\mathbf{q}^{\mathcal{P}}) - \mathbf{K}_{\text d}\mathbf{\dot{q}}^{\mathcal{P}} \\
    \boldsymbol{\tau}^{\mathcal{S}} &=  (\mathbf{J}^{\mathcal{S}})^{T}(\mathbf{J}^{\mathcal{P}})^{-T}\boldsymbol{\tau}^{\mathcal{P}}
\end{align}
\end{subequations}
where the $\textbf{FK}^{\mathcal{P}}$ and $\textbf{IK}^{\mathcal{P}}$ represent the closed chain forward kinematics and inverse kinematics between the joint space and certasian space~\cite{sigma-hopper}. The $\textbf{FK}^{\mathcal{S}}$ and $\textbf{IK}^{\mathcal{S}}$ represent the forward kinematics and inverse kinematics for the template serial joint configuration in the simulation. $\mathbf{J}^{\mathcal{P}}$ and $\mathbf{J}^{\mathcal{S}}$ are the Jacobian matrix for the parallel and serial configuration respectively. $\mathbf{K}_{\text p} = 20$ and $\mathbf{K}_{\text d} = 0.5$ are the parameters of the low-level PD controller.

Since the policy is designed to be agnostic to the parallel and serial joint conversion, no kinematics or torque conversion is needed during the policy deployment on the real parallel hopping robot as shown in Fig.~\ref{frame}.



\section{Simulation Validation}


\begin{figure}[!t]
\centering
\includegraphics[width=0.99\linewidth]{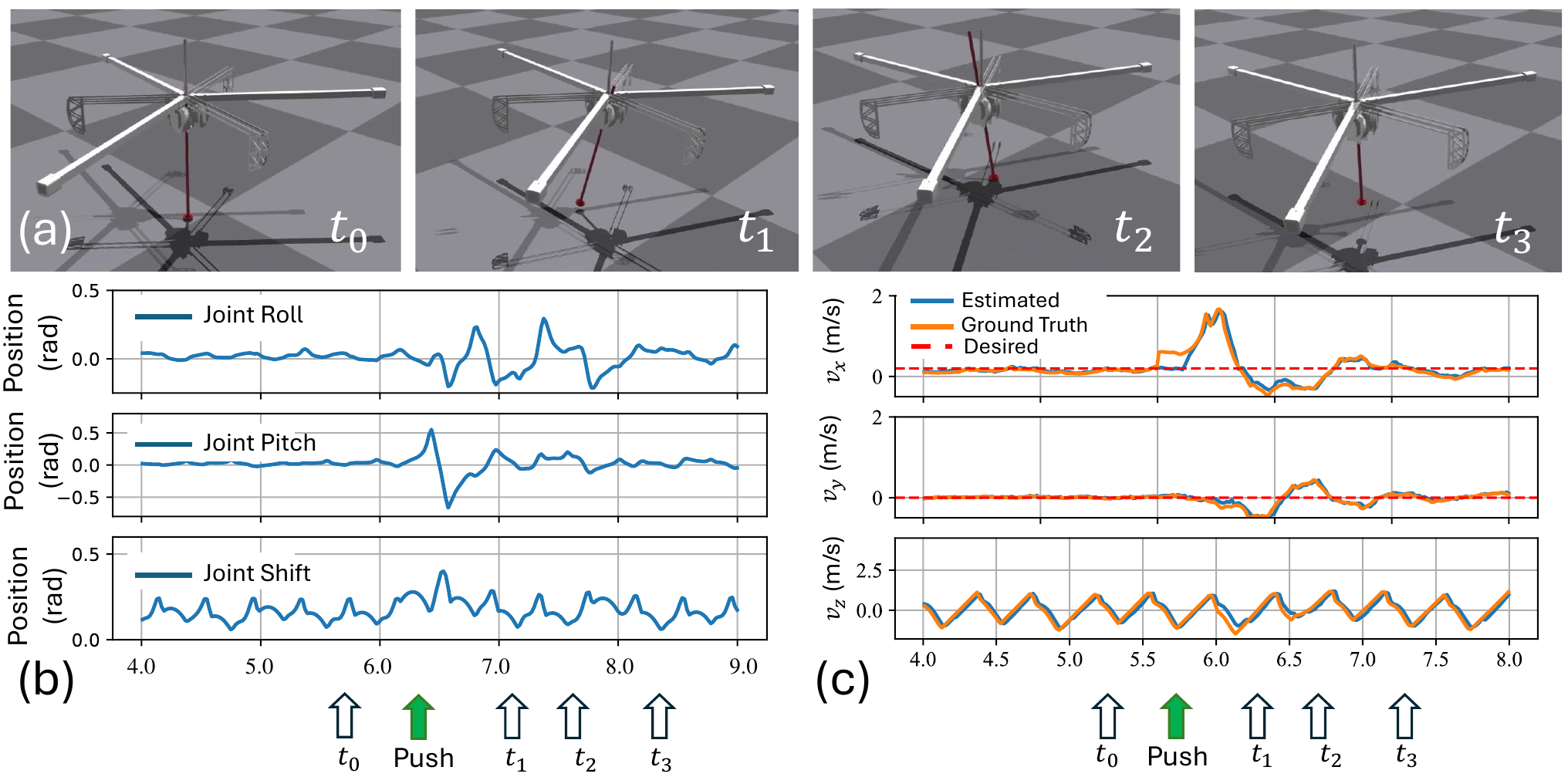}

\caption{Joint position and the estimated robot state obtained from the $\beta$-VAE encoder are shown. 
}
\label{sim-result}
\vspace{-0.5cm}
\end{figure}
In this section, we conduct experiments in simulation to verify several design choices of the control framework in the simulation through an ablation study. We first compare different design choices of the policy latent space and state estimation.

\subsection{Simulation Setup}
As shown in Fig.~\ref{sim-result}, the hopping robot is commanded to jump forward at a speed of $0.2m/s$ with a hopping period of $0.4s$. A $0.4m/s$ perturbation on the robot's base linear velocity is applied at 5.7s. 
After the perturbation occurs, the robot slightly breaks the phase schedule and produces a foot position deviation to maintain balance. The robot fully recovers and hops normally around $2.0s$ after the perturbation happens. Fig.~\ref{sim-result}(b) shows the templated serial joint position value, while Fig.~\ref{sim-result}(c) shows the explicit estimation result of the base linear velocity and the corresponding ground truth value.

\subsection{Ablation Study over Policy Structure}
We compare the hopping result over the performance of velocity tracking given different velocity commands $\dot{\mathbf{q}}^{\text{d}}_{[x,y],t}$ and hopping period $T$. We majorly focus on the design of latent space and state estimation for the policy. Our proposed method and the two baselines are detailed:

\textbf{Proposed (State Estimation + Latent Space)}:
The proposed method builds both a latent representation and explicit estimation of velocity and contact from the history of observations.

\textbf{State Estimation Only}:
Only the state estimation is built with MSE Loss.

\textbf{Latent Space Only}:
Only the latent representation is built with the VAE-style training.
\begin{figure*}[!t]
\centering
\includegraphics[width=0.99\linewidth]{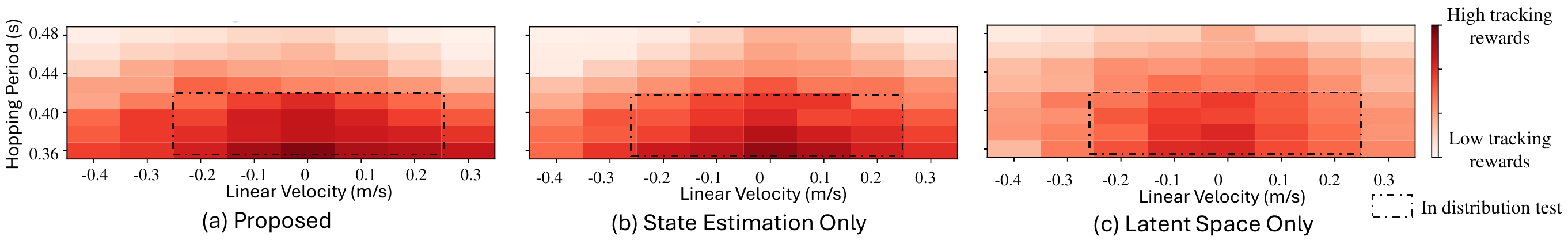}
\caption{Ablation study results over the choice of latent space and explicit estimation.   
}
\label{sim_ablation}

\end{figure*}

As shown in~\ref{sim_ablation}, the darker red means a better tracking performance. The proposed results show an overall better performance over the two baseline methods in both in-distribution and out-of-distribution test. Compared with \textbf{Latent Space Only}(~\ref{sim_ablation} (c)), we conclude that adding explicit base velocity tracking and contact state estimation is crucial to train a velocity tracking hopping policy with long aerial time. Only implicit latent space built by reconstruction is not enough. The comparison between the proposed method and the \textbf{State Estimation Only} shows that adding extra embeddings to allow for implicit latent vector extraction also helps the training. However, when the hopping period becomes longer (longer than 0.44s), the proposed method fails to outperform the \textbf{State Estimation Only}. This may be due to overfitting caused by the addition of a more informative latent space.


\section{Real-World Experiments}
\subsection{Hardware Setup}
\begin{figure}[!t]
\centering
\includegraphics[width=0.85\linewidth]{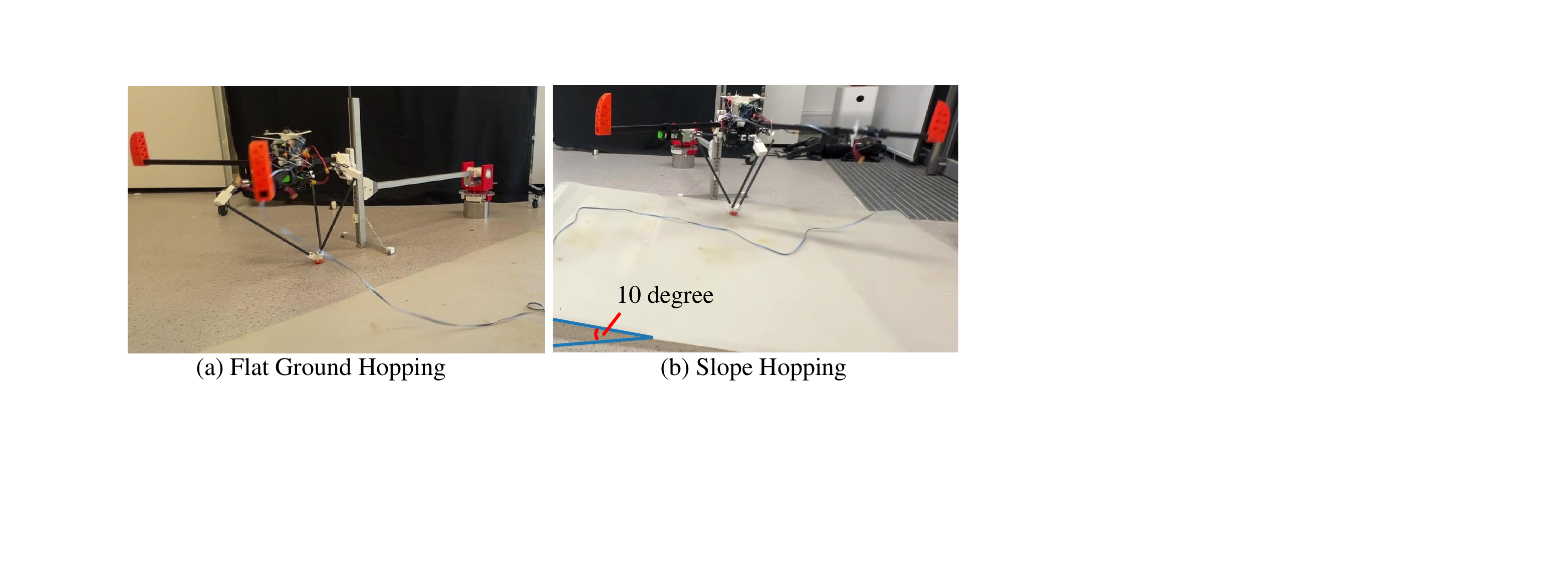}
\caption{Real-world testing scenarios for the hopping robot. (a) Flat ground scene (b) Slope scene.}
\label{real-settingup}

\end{figure}

We conduct the experiments on the real robot in a 2D manner with two scenarios:
\begin{itemize}
    \item \textbf{Flat Ground}: The robot is commanded to hop on flat ground at different commanded velocities as shown in Fig.~\ref{real-settingup}(a).
    \item \textbf{Slope Ground}: A 10-degree slope is placed under the hopper as shown in Fig.~\ref{real-settingup}(b).
\end{itemize}

We design a 2D hopping mode to perform hopping in a 2D plane to testify the hopping performance under different terrains (flat ground and slope). To validate the effectiveness of the proposed learning framework and the sim-to-real transfer method, extensive real world experiments are conducted.
\begin{figure}[!t]
\centering
\includegraphics[width=0.99\linewidth]{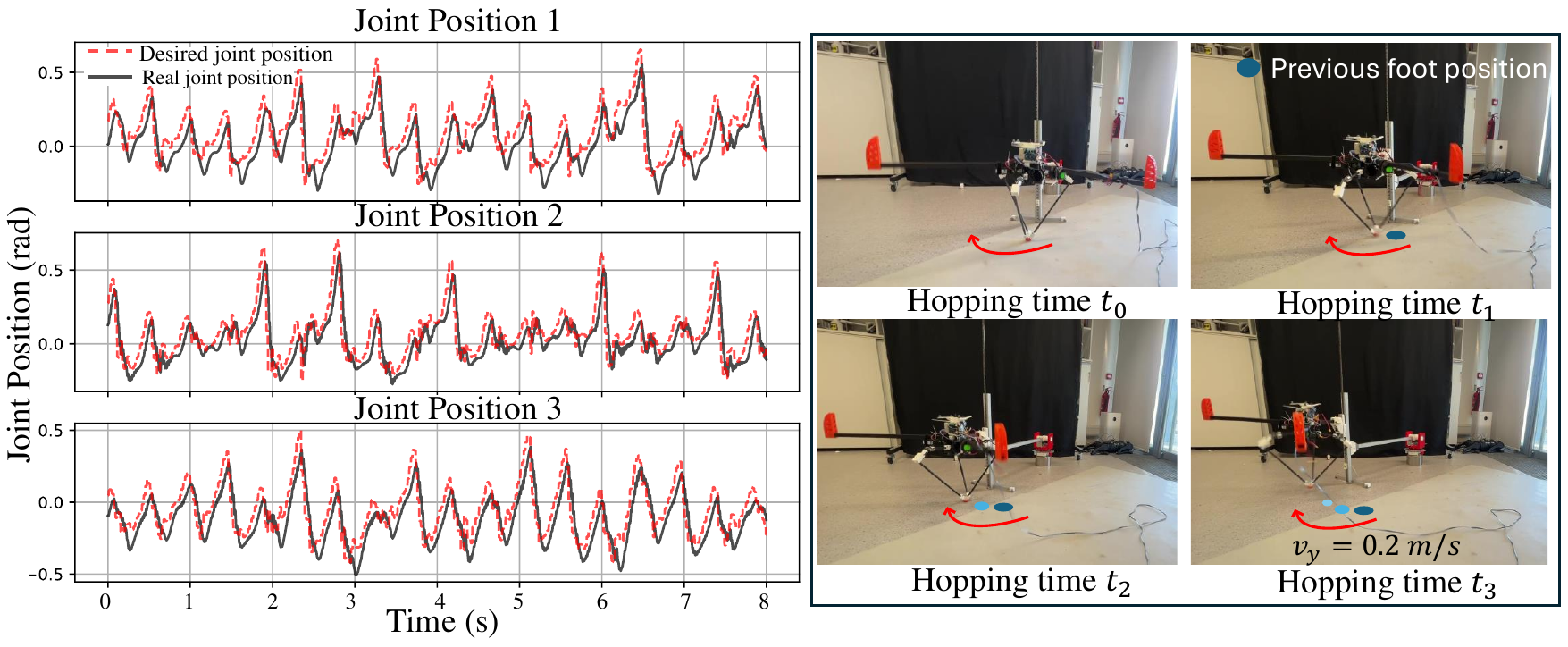}
\vspace{-0.3cm}
\caption{Desired joint positions and real joint positions during hopping with a commanded velocity of $0.2m/s$ are shown. The corresponding snapshots of the hopping robot are presented on the right.}
\label{y_02_joint_space}

\end{figure}
\begin{figure}[!t]
\centering
\includegraphics[width=0.99\linewidth]{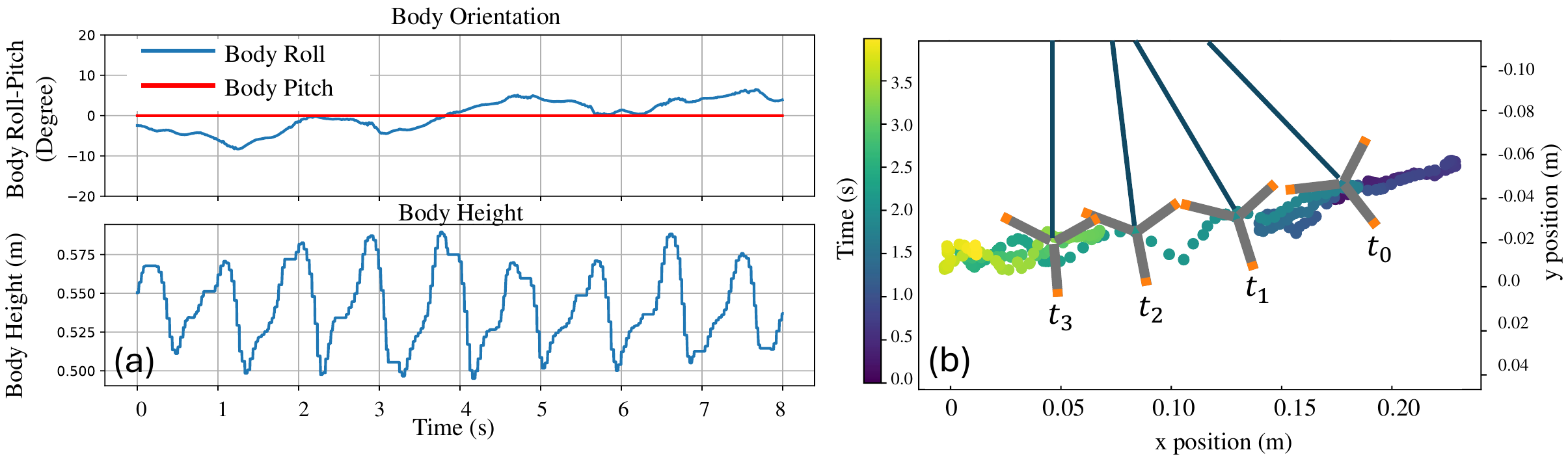}
\vspace{-0.5cm}
\caption{\textbf{Experimental results of continuous hopping motion}. (a) Evaluation of body orientation and body height. (b) Bird-eye view of body trajectory in the x-y plane. The color dots represent the body positions changing with time. Both body height and body position are recorded from a motion capture system.}
\label{base_state}

\end{figure}

\subsection{Continuous Hopping Performance}
We first deploy the trained policy $\pi$ with the  $\beta=0.005$ on the real hopping robot on the flat ground. As shown in Fig.~\ref{y_02_joint_space}, the robot is commanded to hop with a gait period of $0.38s$ with a speed of $0.2m/s$ along the y-axis in the body frame. The speed direction is represented as red arrows in the plot. The outputs ofthe desired joint position with parallel configuration $\mathbf{q}^{\mathcal{P}}$ and the real joint position are plotted. The result shows that the policy trained in simulation with a serial-parallel conversion can be zero-shot transfer to the real world without any fine-tuning or extra conversion. 

During the hopping, the changes in body orientation, body height, and body x-y position are recorded by a motion capture system. As shown in Fig.~\ref{base_state} (a), the roll angle is kept close to zero during the whole hopping motion. The body height changes periodically ($T=0.38s$) according to the designed gait. The global position of the base link is recorded and plotted in Fig.~\ref{base_state}.

\subsection{Comparison on Hopping Controllers}
We also compared our proposed method with a baseline method in the real world. We chose to test different methods on the slope scenario. For the evaluation metrics, we chose \textbf{Surviving Time} and \textbf{Position Tracking Error} to measure the hopping performance. The survival time is measured from the moment hopping begins until the robot either falls over or leaves the safe zone. The position tracking error is determined by averaging the mean square displacement deviation from the starting point.
We conducted each experiment repeatedly 5 times and calculated the average survival time for the metric.
\begin{itemize}
    \item \textbf{SLIP-based controller}~\cite{sigma-hopper}: A model-based controller with a simplified model assumption of the robot is proposed as a baseline controller.
    \item \textbf{RL controller (Proposed)}: We use a reinforcement learning framework with a sim-to-real transfer module to control the robot.
\end{itemize}

\begin{figure}[!t]
\centering
\includegraphics[width=0.99\linewidth]{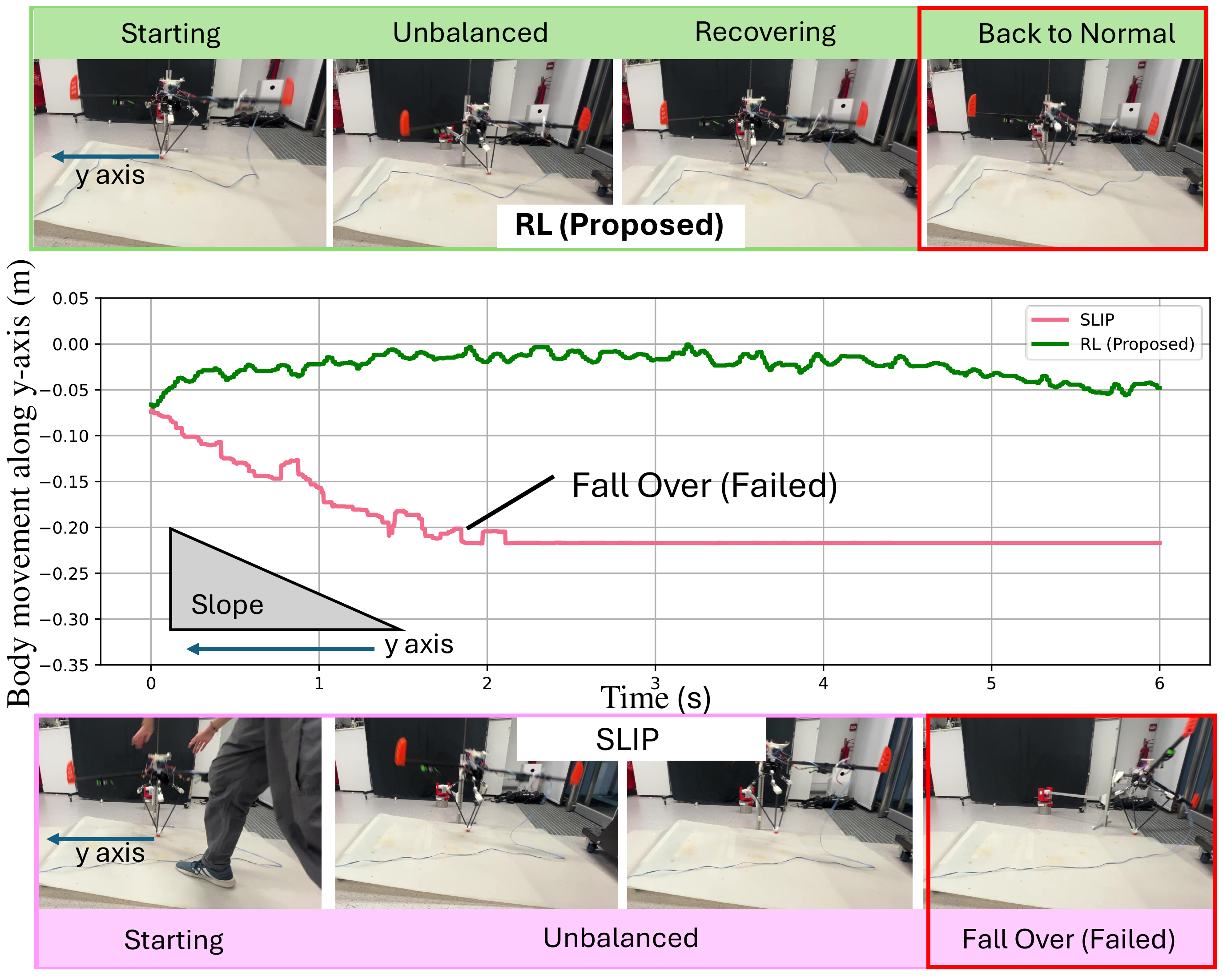}
\caption{Comparison between the proposed method and SLIP based method over the slope terrain. The proposed method successfully recovered from the unbalanced state and quickly rebalanced to a normal hopping state. Instead, the SLIP-based controller failed to regain balance and suffered from a constant drifting toward the negative direction along the y-axis.}
\label{slope_comparsion}

\end{figure}

The comparison results are shown in Table~\ref{tab:controller}. The proposed method largely outperforms the SLIP baseline with both $157\%$ longer surviving time and smaller position tracking errors. This is due to the insufficient assumption of the SLIP based controller under uneven terrains. Instead, the reinforcement learning-based method overcomes uneven terrains through trial and error. We have also plotted the y position of the robot during the hopping motion as shown in Fig.~\ref{slope_comparsion}. The SLIP-based controller constantly drifts towards the downhill direction and finally leads to failure. Instead, the proposed method maintains a relatively stable movement along the y-axis.



\begin{table}
    \centering
    \caption{Comparison of Hopping Performance with 5 Trials}
    \resizebox{0.99\linewidth}{!}{
    \begin{tabular}{ccc}
        \hline
        \textbf{Methods} & \textbf{Surviving Time (s) $\uparrow$}& \textbf{Position Tracking Error $\downarrow$}\\\hline
        SLIP-based controller \cite{sigma-hopper} &   $3.12 \pm 0.48$ & $0.0248 \pm 0.0309$ \\
        RL controller (Proposed) & $\mathbf{8.04 \pm 1.34}$ & $\mathbf{0.0039 \pm 0.0034}$
    \end{tabular}
    }
    \label{tab:controller}
\end{table}

\subsection{Comparsion on Handling Sim-to-Real Gaps}
\begin{figure}[!t]
\centering
\includegraphics[width=0.99\linewidth]{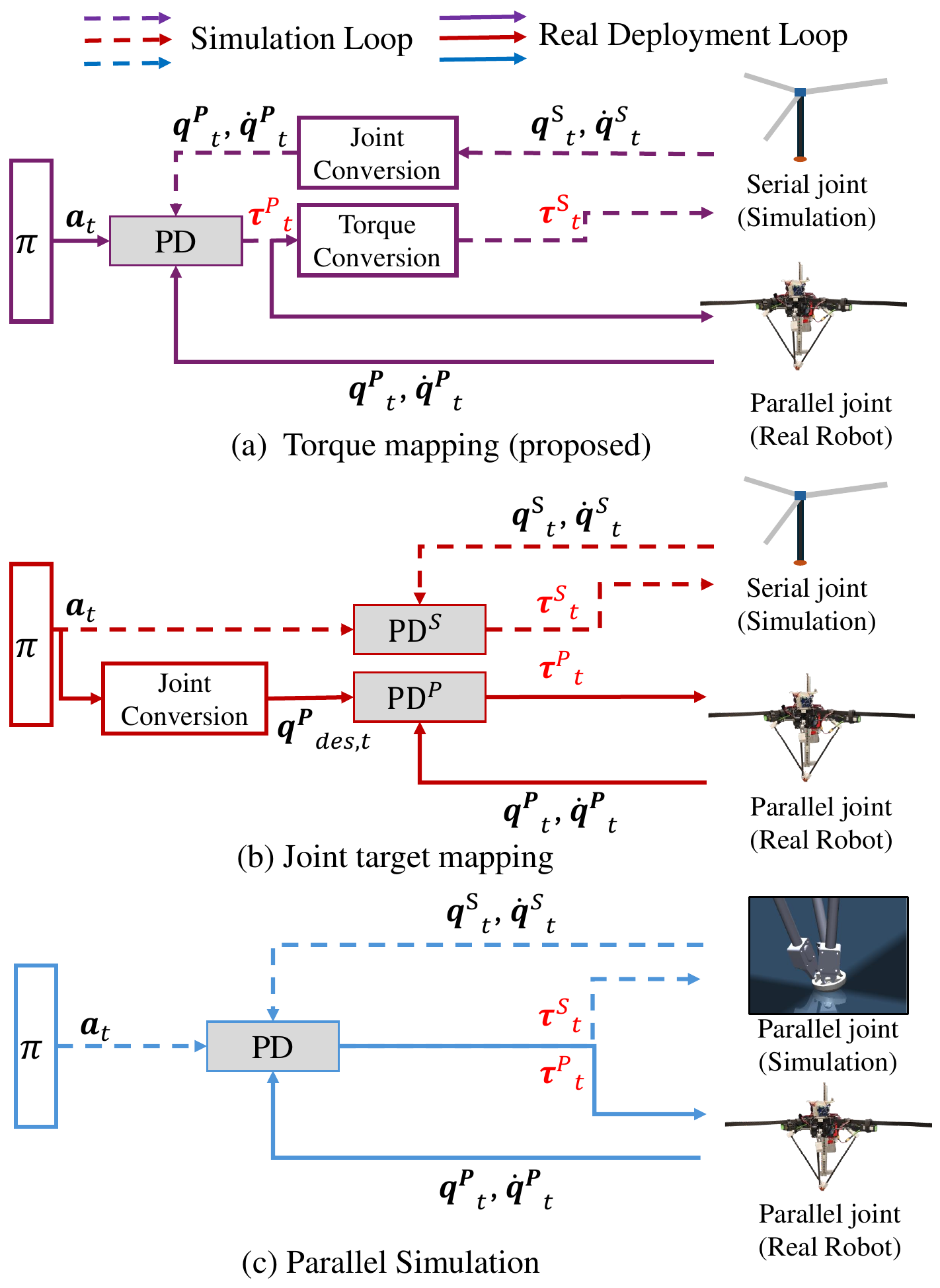}
\caption{ The proposed and several baseline serial-parallel conversion methods in the simulation loop and real-world deployment loop are sketched.}
\label{sim_to_real_comparsion}
\end{figure}

To validate the effectiveness of the proposed sim-to-real conversion methods, we select two baselines that are commonly used when controlling robots with parallel designs or closed chains kinematics. 
\begin{itemize}
    \item \textbf{Torque mapping (proposed)}: We use torque-level mapping through Jacobian during the training process and remove the mapping in the real-world deployment.
    \item \textbf{Joint target mapping}~\cite{gu2024advancing}: We directly conduct a remapping between the policy output joint desired position and the real joint desired position using a kinematics conversion. Then a PD controller converts the tracking error into the real robot joint torque.
    \item \textbf{Parallel simulation}~\cite{li2023robust}: We utilize Mujoco~\cite{todorov2012mujoco} platform, which provides us with a relatively useful 3D parallel mechanism simulation to train the policy and implement sim-to-real transfer.
\end{itemize}
As shown in Fig.~\ref{sim_to_real_comparsion}, we visualize the control loop of both training and real-world deployment for dealing with the sim-to-real transfer. The major differences between these methods lie in the methods of dealing with serial-parallel conversion.

\begin{figure}[!t]
\centering
\includegraphics[width=0.99\linewidth]{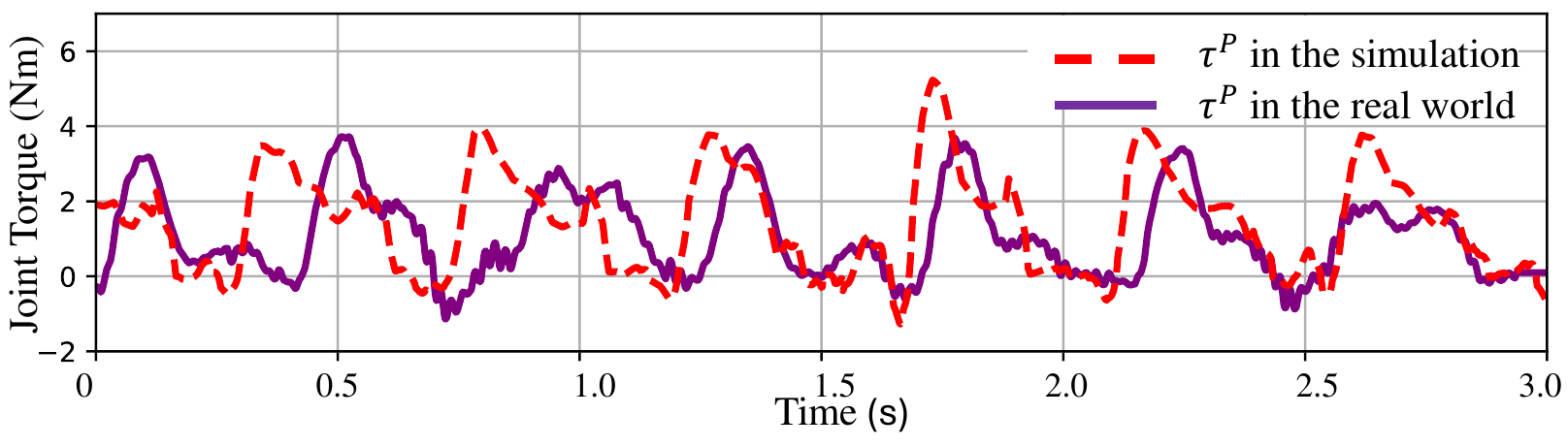}
\caption{The plots of joint torque in the simulation with torque mapping and the real joint torque controlled by the same policy are shown. Similar torque patterns after $1.2 s$ demonstrate the small sim-to-real gap using the proposed torque mapping method.}
\label{torque_mapping_plot}

\end{figure}

To compare the mentioned three methods above, we conduct 2D hopping experiments on flat ground. For the \textbf{Torque mapping (proposed)}, we first visualized the joint torque curve of the same policy under simulation and physical platform deployment in Fig.~\ref{torque_mapping_plot}. The joint torque $\boldsymbol{\tau}^{\mathcal{S}}$ in the simulation is obtained from the torque conversion process depicted in the diagram Fig.~\ref{sim_to_real_comparsion} (a). 
We observe that the torque patterns between the simulation and the real-world deployment are similar. The real hopping robot experienced a slight imbalance during the first $1.2 s$ but regained stability and performed consistent jumps afterwards. After $1.2 s$, the torque patterns in the simulation and the real world aligned closely. This consistency further validates the effectiveness of our proposed torque-level mapping sim-to-real conversion method.

Instead, the $\textbf{Joint target mapping}$ failed to conduct the sim-to-real transfer. We hypothesize that the failure is due to the following reasons: (1) Kinematic-level mapping is insufficient to capture the complex kino-dynamic difference between the serial and parallel joint configuration. (2) The parameter of the $\text{PD}^{\mathcal{S}}$ used during the real-world deployment is unknown and should be empirically selected. This makes sim-to-real even more difficult.

\begin{figure}[!t]
\centering
\includegraphics[width=0.99\linewidth]{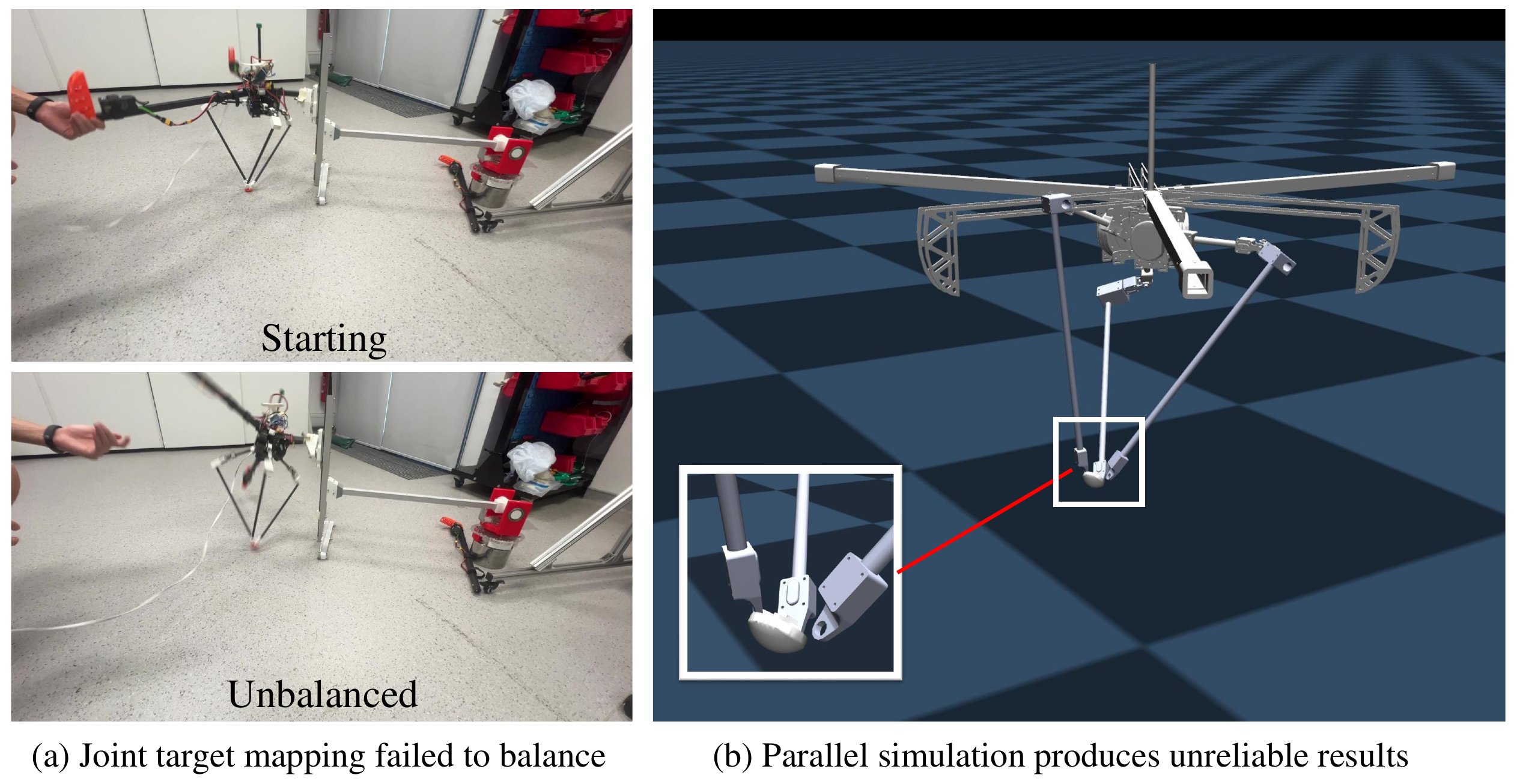}
\caption{The baseline \textbf{Joint target mapping} fails during the sim-to-real transfer. Besides, the \textbf{parallel simulation} in Mujoco fails to provide a reliable simulation result given our 3-RSR leg.}
\label{failure-baseline}

\end{figure}

We also attempted to simulate the parallel mechanism directly in the simulation as shown in~\ref{failure-baseline} (b). However, due to the limitation of the simulator, we failed to build an accurate parallel joint configuration that is stable enough to hop. Different from simulating 2D or 1D closed chain mechanisms~\cite{li2023robust}, the 3D parallel mechanism makes it harder to have an accurate and high-fidelity simulation result. We observed that the simulated leg had a non-unique kinematics mapping from the joint position to the foot position. We also observed that the three joints produced a coupling effect, which made the joint torque affect each other.

\section{Conclusions}
This work focuses on developing a continuous hopping controller for a single-legged robot equipped with a parallel mechanism-based leg. The controller leverages reinforcement learning using an equivalent serial mechanism and torque-level conversion between serial and parallel mechanisms. We hypothesize that our proposed hopping framework can be a general control paradigm when dealing with under-actuated systems. Besides, the proposed sim-to-real conversion method is platform-independent and can be applied to other robots with parallel designs.
In the future, we plan to achieve 3D hopping with the robot and further enhance its motion capabilities in complex terrains.

\bibliographystyle{IEEEtranN}
{\footnotesize \bibliography{bib/example}}
\vfill

\end{document}